\documentclass[10pt,twocolumn,letterpaper]{article}

\usepackage{tabularx}
\usepackage{longtable}
\usepackage{multirow}
\usepackage{afterpage}
\usepackage{wacv}              

\usepackage{graphicx}
\usepackage{amsmath}
\usepackage{amssymb}
\usepackage{enumitem}
\usepackage{booktabs}
\usepackage{comment}
\usepackage{mathtools}

\usepackage[pagebackref,breaklinks,colorlinks]{hyperref}

\usepackage[capitalize]{cleveref}
\crefname{section}{Sec.}{Secs.}
\Crefname{section}{Section}{Sections}
\Crefname{table}{Table}{Tables}
\crefname{table}{Tab.}{Tabs.}

\def\wacvPaperID{1416} 
\def\confName{WACV}
\def\confYear{2025}

\begin{document}

\twocolumn[{%

{ \large
\begin{itemize}[leftmargin=2.5cm, align=parleft, labelsep=2cm, itemsep=4ex,]

\item[\textbf{Citation}]{K. Kokilepersaud, S. Kim, M. Prabhushankar, G. AlRegib, "HEX: Hierarchical Emergence Exploitation in Self-Supervised Algorithms," in \textit{2025 Winter Applications of Computer Vision (WACV), Tuscon, Arizona, 2025.}}

\item[\textbf{Review}]{Date of Acceptance: October 28th 2024}

\item[\textbf{Codes}]{\url{https://github.com/olivesgatech/HEX}}

\item[\textbf{Bib}]  {@inproceedings\{kokilepersaud2025hex,\\
    title=\{HEX: Hierarchical Emergence Exploitation in Self-Supervised Algorithms\},\\
    author=\{Kokilepersaud, Kiran and Kim Seulgi and Prabhushankar, Mohit and AlRegib, Ghassan\},\\
    booktitle=\{Winter Applications of Computer Vision (WACV)\},\\
    year=\{2025\}\}}


\item[\textbf{Contact}]{
\{kpk6, seulgi.kim, mohit.p, alregib\}@gatech.edu\\\url{https://alregib.ece.gatech.edu/}\\}
\end{itemize}

}}]

\title{\ HEX: Hierarchical Emergence Exploitation in Self-Supervised Algorithms}

\author{Kiran Kokilepersaud \and Seulgi Kim \and Mohit Prabhushankar \and Ghassan AlRegib\and
OLIVES at the Center for Signal and Information Processing CSIP\\
School of Electrical and Computer Engineering, Georgia Institute of Technology, Atlanta, GA, USA\\
{\tt\small \{kpk6, seulgi.kim,  mohit.p, alregib\}@gatech.edu}
}
\maketitle

\begin{abstract}
   In this paper, we propose an algorithm that can be used on top of a wide variety of self-supervised (SSL) approaches to take advantage of hierarchical structures that emerge during training. SSL approaches typically work through some invariance term to ensure consistency between similar samples and a regularization term to prevent global dimensional collapse. Dimensional collapse refers to data representations spanning a lower-dimensional subspace. Recent work has demonstrated that the representation space of these algorithms gradually reflects a semantic hierarchical structure as training progresses. Data samples of the same hierarchical grouping tend to exhibit greater dimensional collapse locally compared to the dataset as a whole due to sharing features in common with each other. Ideally, SSL algorithms would take advantage of this hierarchical emergence to have an additional regularization term to account for this local dimensional collapse effect. However, the construction of existing SSL algorithms does not account for this property. To address this, we propose an adaptive algorithm that performs a weighted decomposition of the denominator of the InfoNCE loss into two terms: local hierarchical and global collapse regularization respectively. This decomposition is based on an adaptive threshold that gradually lowers to reflect the emerging hierarchical structure of the representation space throughout training. It is based on an analysis of the cosine similarity distribution of samples in a batch. We demonstrate that this hierarchical emergence exploitation (HEX) approach can be integrated across a wide variety of SSL algorithms. Empirically, we show performance improvements of up to 5.6\% relative improvement over baseline SSL approaches on classification accuracy on Imagenet with 100 epochs of training.  
\end{abstract}

\section{Introduction}
\label{sec: introduction}

\begin{figure}[ht]
\centering
\includegraphics[scale = .25]{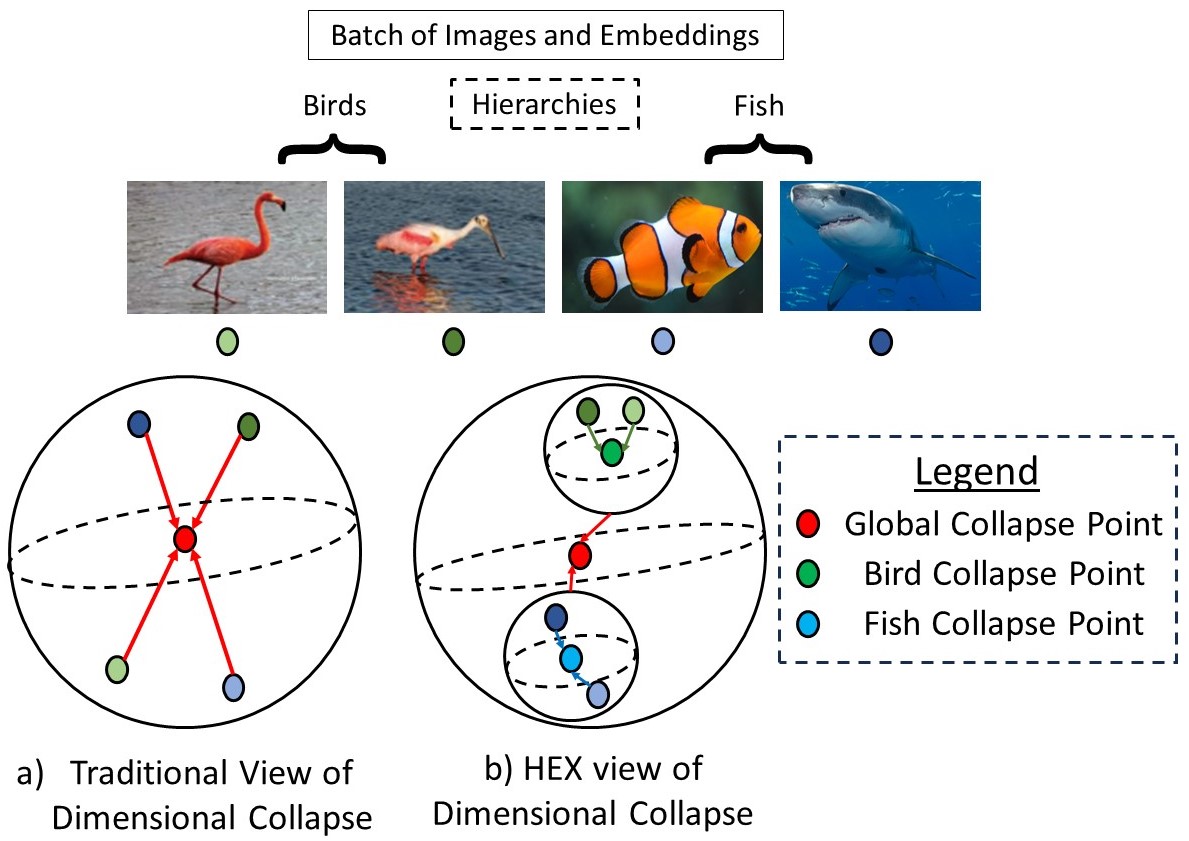}
\caption{a) Traditional view of dimensional collapse is all samples mapping to a single point. b) HEX view of dimensional collapse is that samples first collapse to local hierarchical regions before collapsing globally.}
\label{fig: collapse}
\end{figure}

\begin{figure*}[ht]
\centering
\includegraphics[scale = .2]{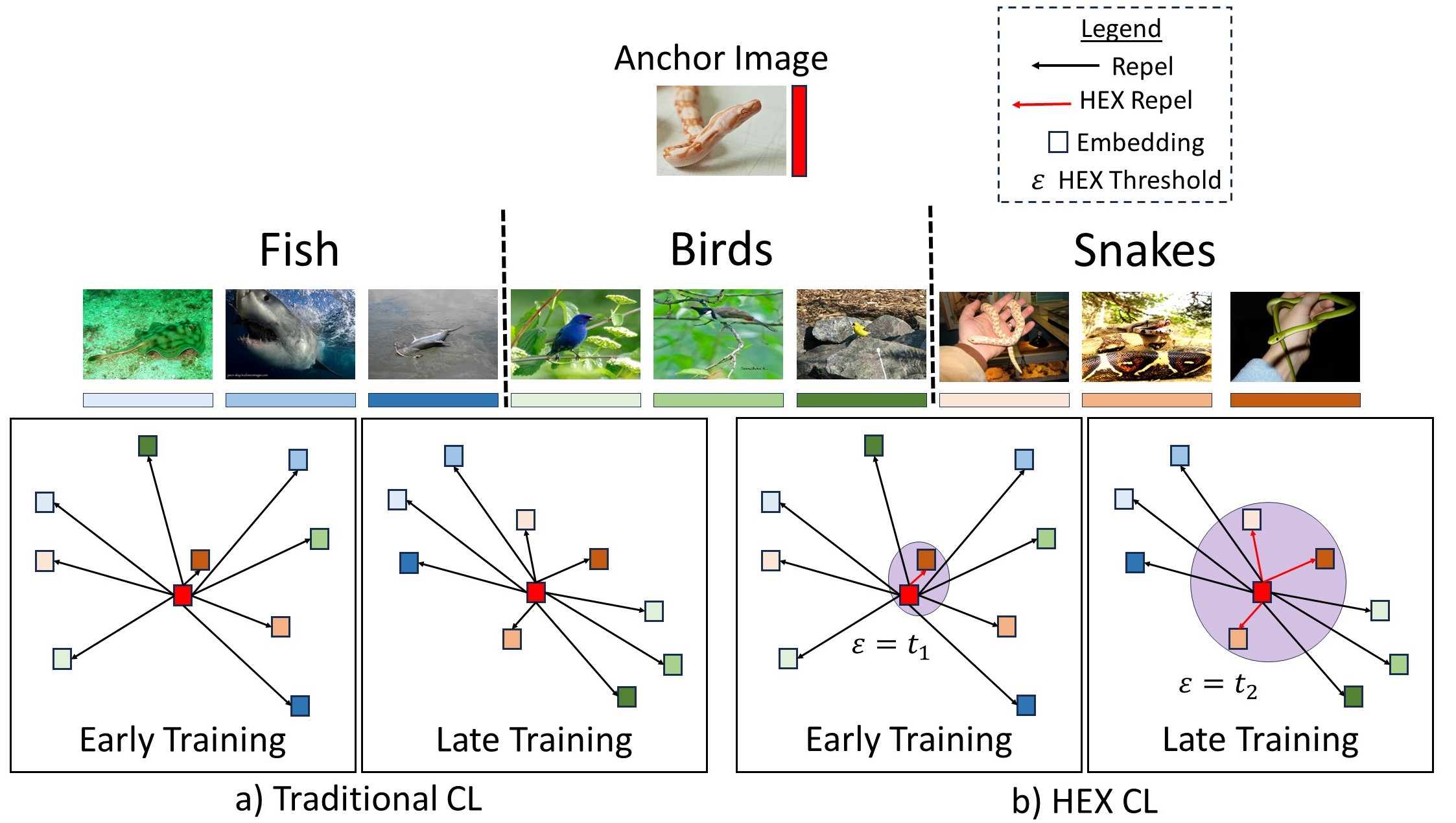}
\caption{a) Traditional contrastive learning algorithms do not adapt their optimization objective to changes in the structure of the representation space as training progresses. b) HEX introduces a custom negative repulsion to samples  above a $\epsilon$ threshold cosine similarity of the anchor image. This $\epsilon$ threshold is varied during training to reflect changes in the structure of the representation space. Early and late training refers to the explicit training time of the algorithm i.e. the number of epochs. }
\label{fig: main}
\end{figure*}

Recent advancements in self-supervised learning (SSL) have resulted in algorithms that approach or  surpass supervised strategies across a wide variety of tasks \cite{chen2020improved,chen2020simple,dwibedi2021little,zbontar2021barlow,bardes2021vicreg}. Generally, all SSL approaches work through a loss that is composed of an invariance term that ensures consistency between similar samples and a regularization term to prevent dimensional collapse. Dimensional collapse \cite{jing2021understanding} refers to representations that only span a lower dimensional subspace. In the most extreme case, this can look like part a) of Figure \ref{fig: collapse} where all embedding vectors map to the same point rather than spanning the complete space. The regularization term of self-supervised algorithms typically addresses this dimensional collapse at a global level whereby all samples of a batch are equally distanced from each other \cite{chen2020simple}. However, this does not account for the emergence of local structures that exist within the representation space. Specifically, recent work has demonstrated that the representation space of self-supervised algorithms gradually reflects a hierarchical structure \cite{ben2023reverse} that becomes more apparent as training progresses. Hierarchically structured representations refers to a representation space where samples are clustered with respect to multiple levels of semantic concepts that they may belong to. For example, in Figure \ref{fig: collapse} we see an example of a flamingo image that belongs to both the flamingo class as well as the higher level bird superclass. Data samples that are members of the same hierarchical grouping typically exhibit features in common with each other such as in the example of Figure \ref{fig: collapse} where birds and fish have similar semantic characteristics. Since these samples share features in common with each other, the representation of these samples can map very close to each other and potentially collapse to the same point \cite{fu2022details}. In contrast to the traditional self-supervised learning paradigm, we view the dimensional collapse phenomenon occurring at a local hierarchical level as well as at a global level as we show in part b) of Figure \ref{fig: collapse} where there exists both global and hierarchy specific collapse points. This perspective presents an opportunity to force the model to recognize more fine-grained differentiating features \cite{prabhushankar2020contrastive} via localized regularization of samples belonging to the same hierarchical grouping. However, the problem is that it is non-trivial to detect the existence of these hierarchical structures and directly integrate them into the associated loss function. This is especially difficult when one considers that different SSL strategies have a wide variety of optimization objectives and architectures. Specifically, SSL algorithms can be taxonomized into dimension-contrastive approaches that optimize for the dimensionality of the space through architectural designs \cite{grill2020bootstrap} and regularization of covariance matrices \cite{bardes2021vicreg} or sample-contrastive approaches that enforces similarity relationships between data points \cite{garrido2022duality}.

In this work, we take advantage of the gradual emergence of hierarchical structures during the training process to introduce a local hierarchical regularization that we call HEX (Hierarchical Emergence Exploitation). We see in Figure \ref{fig: main} a batch of images that roughly exists as part of three different hierarchies: fish, birds, and snakes. Early in training, these points are randomly dispersed throughout the representation space. However, as training progresses, members of the same hierarchical grouping become more aligned with each other. In part a), we see that traditional contrastive learning does not account for this gradual hierarchical organization and optimizes the batch the same way at both the start and end of training. However, in contrast to part a) we take advantage of this hierarchical emergence to introduce HEX repulsion in part b) of Figure \ref{fig: main}. Essentially, we identify samples that are above an $\epsilon$ threshold of cosine similarity with the anchor image as members of the same hierarchical grouping as the anchor image. Samples that fit this threshold criteria are visually shown to lie within the colored region in part b) of Figure \ref{fig: main}. We then assign an additional weighting term to these samples within the denominator of the InfoNCE \cite{oord2018representation} loss that effectively decomposes it from the summation of other samples in the batch and functions as an addtional localized regularization to counter the local collapse effect. This $\epsilon$ threshold is gradually relaxed during the training process to reflect greater confidence in the hierarchical alignment of samples later in training. We show how this $\epsilon$ parameter can be chosen manually and adaptively. Additionally, the HEX loss naturally works on top of all sample contrastive approaches due to regularizing sample-wise relationships, but we also show that it can be used as a regularization term on top of dimension contrastive strategies as well.

In summary, the contributions of our work include:
\begin{enumerate}[topsep=0pt,itemsep=-1ex,partopsep=1ex,parsep=1ex]
    \item We propose a self-supervised methodology based on cosine similarity distributions to identify the existence of local hierarchical groupings of images within the representation space. 
    \item We show how these identified hierarchical groupings can be integrated within the InfoNCE loss to introduce a local hierarchical regularization into the training process for both sample and dimension contrastive strategies. 
    \item We introduce an $\epsilon$ threshold term during the training process that can be tuned adaptively or manually to reflect the growing hierarchical organization of the representation space. 
    \item We show performance improvements of our approach across a wide variety of relevant classification scenarios such as fine-grained recognition, large scale evaluations, and across tasks of varying diversity. 
\end{enumerate}

\section{Related Works}
\label{sec: related_works}
\subsection{Self Superivsed Learning} The literature on self-supervised learning is divided between sample and dimension-contrastive approaches \cite{garrido2022duality}. Both approaches have an invariance term that enforces consistency between similar samples and a regularization term to prevent dimensional collapse. However, the specific optimization details differs between each SSL algorithm. Sample contrastive approaches such as \cite{chen2020simple,dwibedi2021little} involve building an embedding space where similar pairs of images (positives) project closer together and dissimilar pairs of images (negatives) project apart. Sample contrastive approaches differ in the manner in which these positive and negative sets are defined. Other approaches include computing a weighting distribution on negative samples \cite{robinson2020contrastive}, adding a memory queue for additional negatives \cite{chen2020improved}, and addressing the problem of false negatives \cite{huynh2022boosting}. Application specific approaches have also defined positive and negative distributions based on the structure of specific types of data modalities. This includes the medical setting \cite{kokilepersaud2023clinically}, seismology applications \cite{kokilepersaud2022volumetric}, or within a fisheye camera object detection scenarios \cite{kokilepersaud2023exploiting}. Dimension contrastive approaches work by exploiting some property of the representation space that results in feature decorrelation. These methods include approaches that use knowledge distillation such as \cite{grill2020bootstrap} and \cite{caron2021emerging}. Other ideas include information maximization strategies that operate on empirical covariance matrices such as \cite{bardes2021vicreg} and \cite{zbontar2021barlow}. Overall, all these strategies prevent dimensional collapse at a global level while not having an explicit mechanism to counter local dimensional collapse between samples of the same hierarchical grouping. We address this gap through our additional regularization term during training. Additionally, our work has the added benefit of being usable on top of both sample and dimension contrastive strategies.

\subsection{Hierarchical Representations}
Other works have defined hierarchy within a representation learning context. However, this notion of hierarchy oftentimes does not align with the specific setting that we address in this paper. For example, there have been a variety of hierarchical representation learning techniques within the context of supervised learning scenarios \cite{khosla2020supervised} where exact knowledge of the nature of the semantic hierarchy is known. This includes works that form representations based on a supervised multi-label setting such as \cite{zhang2022use} and \cite{sajedi2023end} where a loss is introduced to address the multitude of different label types. Other work introduces hierarchical label sets to supplement the supervised contrastive loss such as the soft label concept in \cite{feng2023maskcon} or the explicit label hierarchy in \cite{lian2024learning}. Hierarchies have also been defined with respect to the nature of the data augmentation such as in \cite{zhang2022rethinking}. Furthermore, works have defined hierarchies with respect to the construction of the task itself as in the case of action recognition \cite{dong2023hierarchical}, 3D segmentation \cite{ying2024omniseg3d}, or within the context of word embedding vectors\cite{hu2022hiure,li2022hiclre}. Our work differs from all these in that we estimate hierarchies that emerge from the representation space within an unsupervised context. Other works implicitly describes hierarchical relationships through the notion of the need for a localized version of the contrastive loss. However, the notion of localized contrastive learning is defined with respect to the application space such as volumetric positions in medical data \cite{chaitanya2020contrastive}, spatio-temporal relationships in video data \cite{zeng2021contrastive}, and local regions for semantic segmentation and object detection tasks \cite{tang2023semantic,chen2023lpcl}. Other work defines localized contrastive learning with respect to features at different layers of a network \cite{xiong2020loco,fournier2023preventing} or analyzing the nature of spectral embeddings \cite{balestriero2022contrastive}. Our work differs due to explicitly introducing a general regularization term to counter the localized dimensional collapse effect that can be used across a wide variety of SSL objectives.

\section{Hierarchy Analysis}
\label{sec: hierarchical_analysis}

In this section, we aim to address the following questions: 1) How can we motivate the need for additional hierarchical regularization during training? 2) What statistical analyses can we perform to estimate samples within a batch that belong to the same hierarchical grouping?

\subsection{Analytical Setup} To perform this hierarchical analysis, we make use of the Cifar-100 dataset \cite{krizhevsky2009learning} that is constructed with a natural hierarchical structure. Specifically, the dataset is composed of 100 classes with each class having an equal number of 500 images. Every set of 5 classes has its own superclass that reflects classes that share semantic features in common. This results in 20 superclasses. The test set is composed in a similar manner with every class having 100 inference samples and the same hierarchical relationships as the training set. We train the NNCLR \cite{dwibedi2021little} method with a ResNet-50 architecture $f(.)$ and an associated projection head $g(.)$ on Cifar-100 for 450 epochs. A model checkpoint is saved every 50 epochs to produce a set of models $m_e \in M$. On each training checkpoint, we pass in the test set sample $x_i \in X$ that has its ground truth label $y_i \in Y$ as well as its superclass label $y_{si} \in Y_s$. $g(f(x_i))$ produces an embedding of the image $z_i$ while $f(x_i)$ produces a representation of the image $r_i$.

\begin{figure}[ht]
\centering
\includegraphics[scale = .45]{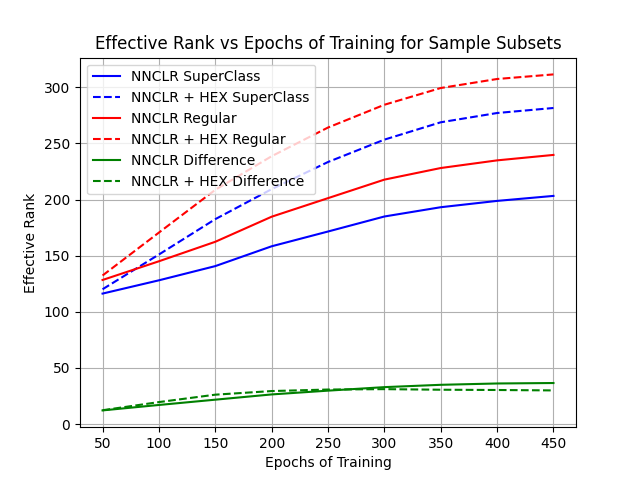}
\caption{Effective rank of 500 sample subsets that are sampled from only superclass samples and random samples across the Cifar-100 test set from a model trained with NNCLR alone and NNCLR + HEX. The effective rank was calulated using the RankMe metric.}
\label{fig: rank}

\end{figure}

\subsection{Hierarchical Collapse} 
To motivate the need for additional hierarchical regularization in the SSL loss function, we analyze the local dimensional collapse property of samples within the same superclass in Figure \ref{fig: collapse}. In this figure, we sample 20 subsets of 500 images for each model $m_e$ at different checkpoints of training. These subsets are varied based on whether all the samples are taken from the same superclass or randomly sampled globally from the test set.  For each subset, we produce a subset representation matrix $R_S$ by passing the images into the model $f(.)$. We then compute the effective rank of this matrix through the RankMe metric \cite{garrido2023rankme} that acts as a smooth measure of the overall dimensionality of the space. For each of the subsets generated, we compute this metric and take the average of all the subsets computed on each epoch. We repeat this process for both the NNCLR \cite{dwibedi2021little} SSL methodology alone as well as with our introduced HEX regularization. 

On NNCLR alone, we see that this effective rank measure is greater for samples drawn globally from the test set compared to samples drawn from the same superclass. Additionally, the difference in this effective rank measure increases between these two curves as training progresses. This indicates that samples from the same superclass exhibits a local form of collapse that is greater than that of samples drawn globally from the test set. Additionally, the difference in the collapse of superclass samples compared to regular samples increases as training progresses. Hence, this illustrates that these samples need additional local hierarchical specific regularization in order to counter this localized collapse. When we introduce our additional HEX regularization on top of NNCLR, we still observe greater collapse from superclass samples compared to regular samples, but the overall dimensionality of both spaces is significantly higher than for NNCLR alone. Additionally, the difference in dimensionality between the superclass and regular subsets initially increases, but begins to decrease as training progresses. All of this indicates that the additional HEX regularization is able to counter dimensional collapse at both local hierarchical and global levels.
\subsection{Hierarchy Estimation}
\begin{figure*}[ht]
\centering
\includegraphics[scale = .3]{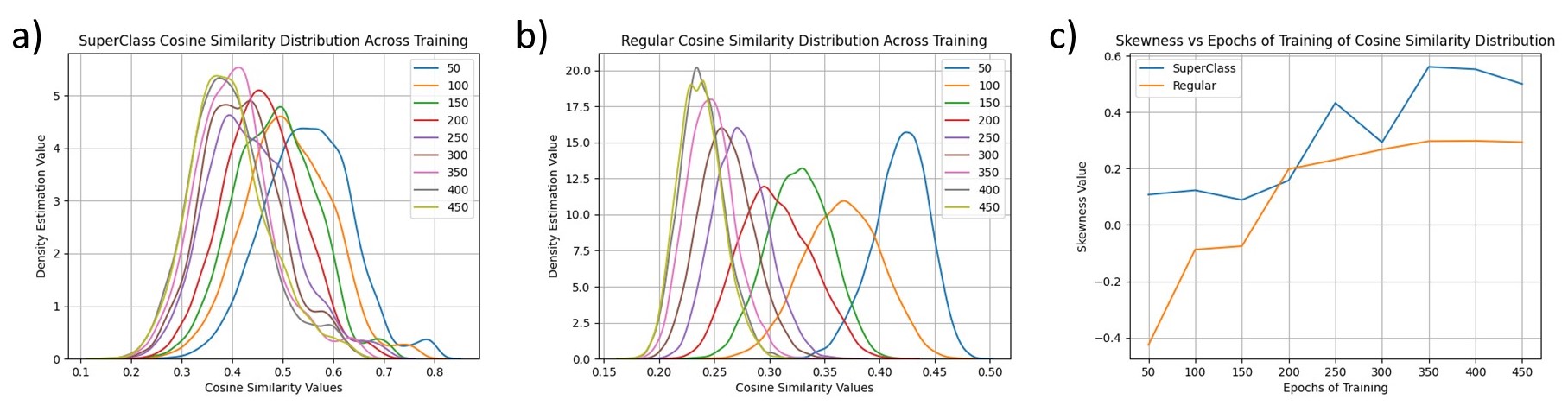}
\caption{a) and b) are plots of the cosine similarity distribution at different epochs of training for the SuperClass samples and Regular samples respectively on the Cifar-100 test set. c) This is a plot of the skew of the cosine similarity distribution for both superclass and regular samples across training.}
\label{fig: distributions}
\end{figure*}

We also want to assess whether there is a statistic to effectively estimate samples that exist within the same hierarchical grouping. With a suitable statistic, we can then effectively decompose samples into hierarchical and regular samples without needing access to the ground truth hierarchy organization of the dataset. Additionally, this statistic should become more separable between hierarchically grouped samples and regular samples as training progresses to reflect the growing hierarchical organization of the representation space. We show these trends in Figure \ref{fig: distributions}. In parts a) and b), we generate the average cosine similarity distribution for samples with the same superclass label as each anchor image and differing superclass labels that we refer to as regular samples. We compute these distributions at every 50 epochs of training for the NNCLR method. We observe that on average the distributions of cosine similarity for samples of the superclass are skewed towards higher values than that of regular samples in part a) of Figure \ref{fig: distributions}. Additionally, the right hand tails of the superclass distributions reach values that are separable from that of the cosine distributions of regular samples found in part b) of Figure \ref{fig: distributions}. This is further evidenced in part c) of Figure \ref{fig: distributions} where the superclass cosine similarity distributions exhibit greater skew towards larger values as training progresses. This illustrates that sampling from the right hand tails of the cosine similarity distributions can effectively identify samples that belong to the same hierarchical grouping as the anchor image.

\section{Methodology}
\label{sec: methodology}

The standard sample contrastive learning framework involves an encoder network $f(\cdot)$ along with a projection network $g(.)$. The objective of the contrastive loss is to create a positive set of samples that should be projected close to each other and a negative set of samples that should be projected apart from the positive set. The training data is composed of unlabeled samples $x_i \in I$ where $I$ is the set of all images. During training, a random augmentation is applied to each image in a batch to produce a corresponding set of augmented images for each anchor image of the form $x_{j(i)}$ which constitutes the positive set. Each image is passed through $f(\cdot)$ to produce a representation $r_i$ and then passed through $g(\cdot)$ to produce an embedding $z_i$ and its corresponding augmented sample embedding $z_{j(i)}$. The rest of the samples in the batch form the negative set that are members of the set $a \in A$ with associated embeddings $z_a$. Additionally, dot products of embeddings are normalized by an associated temperature parameter $\tau$. This results in the standard form of the InfoNCE loss that looks like:

\begin{equation}
  L_{NCE} = -\sum_{x_i\in{I}} log\frac{exp(z_{i}\cdot z_{j(i)}/\tau)}{\sum_{a\in{A(i)}}exp(z_{i}\cdot z_{a}/\tau)}
  \label{eq: standard}
\end{equation}

By applying logarithmic rules to the above equation, we can decompose this loss into the summation of an invariance term and a regularization term. The decomposed form can be expressed as:
    $\sum_{x_i\in{I}}(-z_{i}\cdot z_{j(i)}/\tau + log({\sum_{a\in{A(i)}}exp(z_{i}\cdot z_{a}/\tau)}))
    \label{eq: decomposed}$
Equation \ref{eq: decomposed} is minimized when the left hand invariance term has a maximal dot product and the right hand regularization term has a minimal dot product. This regularization term is the mechanism that prevents global dimensional collapse. However,  the regularization term does not account for localized semantic relationships that exist from the emergence of hierarchical relationships. Samples that are members of the same hierarchical grouping may require an additional re-weighting in order to explicitly counter the local dimensional collapse effect of these samples with respect to the anchor image.

\begin{figure*}[ht]
\centering
\includegraphics[scale = .35]{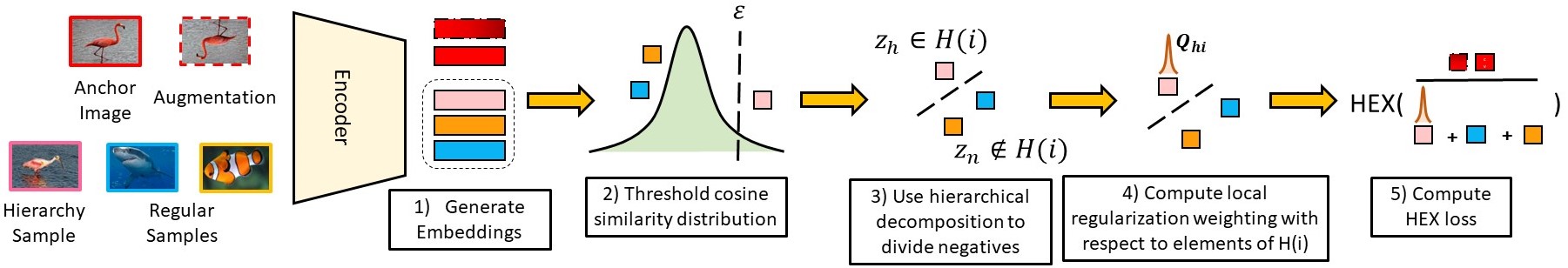}
\caption{This shows the overall process by which the HEX loss is computed. Embeddings are divided by a cosine similarity threshold $\epsilon$ into a hierarchy $H(i)$ and regular subgroup. Hierarchy samples receive additional regularization with the $Q_{hi}$ function and this construction is integrated into the HEX loss.}
\label{fig: method}
\end{figure*}

To address this, our goal is to decompose the regularization term from $log(\sum_{a\in{A(i)}}exp(z_{i}\cdot z_{a}/\tau))$ to $log( Q_{hi}(\sum_{h\in{H(i)}}exp(z_{i}\cdot z_{h}/\tau)) + \sum_{n\notin{H(i)}}exp(z_{i}\cdot z_{n}/\tau)) $ where $H(i)$ is a subset of $A(i)$ that represents members of the same hierarchical grouping as $z_i$ in the batch. We introduce a threshold parameter $\epsilon$ during training to identify samples above a specific cosine similarity value as belonging to $H(i)$. $Q_{hi}$ represents a weighting function specific to the current batch. Its goal is to place an additional weight on samples in $H(i)$ such that an additional localized regularization takes place. There are a variety of functions that could be used to appropriately scale the weighting of these hierarchical negatives. In this case, we use a function similar to the work of \cite{robinson2020contrastive}, but the difference is that its averaging only happens with respect to the elements belonging to $H(i)$ and no additional $\beta$ or $\tau^{+}$ parameter is used. The overall construction of this function $Q_{hi}$ involves computing the sum of the multiplication between cosine similarity values and the exponential of the cosine similarity values for the sample of interest and its hierarchical negatives. This is then normalized by the mean of the exponential of cosine similarity values between the sample of interest and its hierarchical negatives. This term is then added to the exponential of the cosine similarities of the positive set scaled by the batch size $N$ and a temperature parameter $\tau$ that is set to .1 for all experiments. This operation is done at a batch level; however, for a single sample $z_i$ and its hierarchical negatives $z_h$ this $Q_{hi}$ operation can be approximated as: 

\begin{equation}
    \frac{\frac{\sum_{h\in{H(i)}}exp(z_{i}\cdot z_{h}/\tau) (z_{i}\cdot z_{h}/\tau))}{\frac{1}{N}\sum_{h\in{H(i)}}exp(z_{i}\cdot z_{h}/\tau)} -  N\tau exp(z_{i}\cdot z_{j}/\tau)}{1-\tau}
\end{equation}

This process can be visualized in Figure \ref{fig: method}. We refer to this new form of the $L_{NCE}$ loss as $L_{HEX}$. In general, the construction of $L_{HEX}$ can be used on top of any SSL algorithm that makes use of some form of $L_{NCE}$ such as in \cite{chen2020simple,dwibedi2021little,estepa2023all4one}. However, $L_{HEX}$ can also be used as a regularization term as part of a combined loss with a dimensional contrastive approach. We refer to the loss of dimensional contrastive methods as $L_{dim}$ and the resultant combined loss as $L_{total} = \alpha L_{HEX} + (1-\alpha) L_{dim}$ where $\alpha$ weights the contribution of both losses. 

 In Figure \ref{fig: method} we show the process by which the HEX loss is computed, but we did not go into detail about how the hierarchical decomposition takes place. Ideally, the labels that represent the hierarchical structure of the dataset would be available as demonstrated by the analysis in the previous section with the Cifar-100 dataset. In order to approximate these hierarchical groupings in an unsupervised manner, we take advantage of the analysis of the previous section that shows that the tails of the cosine similarity distribution of samples within the same hierarchical grouping are separable from those of regular samples. Therefore, the decomposition can be approximated by choosing an appropriate threshold $\epsilon$ with samples above this threshold being members of $h\in H(i)$ while samples below this threshold are $n\notin H(i)$. Additionally, this threshold changes during training to reflect the shifting cosine similarity distribution as well as the growing confidence that samples with higher cosine distributions will be members of the same hierarchical grouping. 
Empirically, we introduce a few different strategies to choose this threshold parameter. The first is a manual strategy that is similar to a learning rate scheduler. This involves setting the threshold initially high and then lowering it every $N$ number of epochs by a fixed amount in a step-wise fashion. In our experiments, the step-wise function threshold is set to .9 initially and is lowered by .1 every 100 epochs or every 25 epochs in the case of the 100 epochs of training for our ImageNet experiments. We also introduce an adaptive thresholding strategy that does not require manually setting parameters at the start of training. To do this, we compute the overall cosine similarity distribution for each anchor image in a batch and find the value that corresponds to two standard deviations above the mean cosine similarity for each individual batch. This value then serves as the threshold to decompose samples into the hierarchical and regular subsets for the regularization term.

\section{Results}
\label{sec: results}

\subsection{Experimental Overview}

We make use of the solo-learn codebase \cite{da2022solo} and use their default method specific parameters for State-of-the-Art (SOTA) comparisons. This includes parameters such as temperature, learning rate, learning rate scheduler, weight decay, augmentation scheme, and optimizer. Parameters that were set constant for all experiments besides ImageNet and iNat21 include a batch size of 256, usage of the ResNet-50 architecture\cite{he2016deep}, and 400 epochs of training time. Further details related to training can be found in Section \ref{sec: train_details}. The results shown are based on either the online linear evaluation or fine-tuning accuracy protocols. Linear evaluation refers to the setting where a linear layer is trained separately with the cross-entropy loss from the backbone model while the SSL algorithm is training. Fine-tuning refers to freezing the weights of the backbone network after SSL training and appending a linear layer that is trained for 100 epochs with an SGD optimizer, .3 learning rate, weight decay of 1e-5, and decay steps at epoch 60 and 80 for all experiments. Each table or figure will specify which evaluation protocol is used, but empirically performance differences are negligible between each. Experiments run on ImageNet \cite{deng2009imagenet} or iNat21 \cite{van2018inaturalist} received 100 epochs of training with a batch size of 256 and maintained the same hyperparameter choices as in the original papers for SimCLR \cite{chen2020simple} and NNCLR \cite{dwibedi2021little}. In each table, a method with + HEX refers to the baseline method receiving the additional HEX regularization during training. Additionally, the ``Type" column refers to the strategy used to set the $\epsilon$ threshold as training progresses as described in Section \ref{sec: methodology}. 

\begin{table}[]
\centering
\small
\resizebox{\columnwidth}{!}{
\begin{tabular}{cccccc}
              &          & \multicolumn{2}{c}{Cifar-100}   & \multicolumn{2}{c}{Imagenet-100} \\ \cline{3-6} 
Method        & Type     & Top-1          & Top-5          & Top-1           & Top-5          \\ \hline
SimSiam    \cite{chen2021exploring}   & N/A      &   65.90             &       90.36         &  80.26               &      95.68          \\ 
Moco v2  \cite{chen2020improved}     & N/A      & 71.01          & 92.67          &      85.18          &        97.02       \\
BYOL    \cite{grill2020bootstrap}      & N/A      & 71.72          & 92.69          &      82.00           &      95.28          \\ \hline
VicReg   \cite{bardes2021vicreg}     & N/A      & 71.36          & 92.45          &     83.84            &      96.56          \\

VicReg  + HEX ($\alpha = .5$)     & Ada      & \textbf{71.88}          & \textbf{92.58}          &     \textbf{84.16}            &       \textbf{96.62}         \\ \hline
Barlow \cite{zbontar2021barlow}  & N/A      & 70.84          & 92.30          &      \textbf{84.70}          &          96.72      \\ 

Barlow + HEX ($\alpha = .5$)  & Ada      & \textbf{71.94}         & \textbf{92.91}         &       \textbf{84.70}          &        \textbf{96.88}        \\\hline

SimCLR   \cite{chen2020simple}     & N/A      & 64.46          & 88.59          & 80.30           & 95.70          \\
SimCLR + HEX  & Ada & \textbf{67.56} & \textbf{90.02} &  \textbf{81.78} & \textbf{95.78}  \\
SimCLR + HEX  & Step & 66.46 & 89.13 & -  & - \\
SimCLR + HEX  & Sup & 69.16 & 90.46 & -  & - \\\hline
NNCLR  \cite{dwibedi2021little}       & N/A      & 70.72          & 92.62          &       83.84          &         96.4       \\
NNCLR + HEX   & Ada     & 71.55 & \textbf{92.75} &      \textbf{ 84.34}         &      \textbf{96.57}           \\
NNCLR + HEX   & Step     & \textbf{71.76} & 92.66 &   -            &     -           \\

NNCLR + HEX   & Sup     & 74.04 & 93.74 &   -            &     -           \\\hline
All4One  \cite{estepa2023all4one}     & N/A      & 72.48          & 93.07          &    85.88             &     97.10           \\
All4One + HEX & Ada     & 72.88 & 93.50 &        \textbf{86.00}         &     \textbf{97.24}           \\
All4One + HEX & Step     & \textbf{73.34} & \textbf{93.70} &    -             &       -         \\
All4One + HEX & Sup     & 73.82 & 93.85 &    -             &       -         \\\hline
\end{tabular}
}
\caption{Performance of HEX on top of existing SOTA approaches and in comparison with others for the Cifar-100 and Imagenet-100 datasets within a linear evaluation setting. Ada stands for the adaptive version of our method and Step is the Stepwise decrease version. For Cifar-100 we show results on both decomposition types, while for Imagenet-100 we show the adaptive version. Sup refers to the idealized setting where the loss has full access to the superclass labels during the decomposition. We bold the highest performance amongst the fully unsupervised strategies in each section. For VicReg and Barlow Twins, we report the performance under the condition of equal weighting with the HEX loss or $\alpha = .5$.}
\label{tab:complete}
\end{table}

\subsection{SOTA Analysis} In Table \ref{tab:complete} we show a complete performance comparison of our HEX loss across the Cifar-100 and Imagenet-100 datasets for a variety of different state of the art algorithms. Generally speaking, the application of our HEX loss improves performance over the regular self-supervised approach in all cases. However, there are some interesting trends to observe within this table. The first is the case where the decomposition of our HEX loss has full access to suprclass labels during training which is denoted by the `` Sup" type in our table. We observe that ground truth access to the superclass labels significantly outperforms the adaptive and stepwise approximations for all methods where it is applied. This indicates that a good decomposition strategy is one that is effectively able to approximate the higher order hierarchy labels during training. This may also partially explain differences in performance improvements for the stepwise and adaptive strategies between different methods. For example, for SimCLR the adaptive HEX strategy performs significantly better on Cifar-100, while  for NNCLR and All4One the stepwise method is better in terms of top-1 accuracy. This is because our method relies on setting an appropriate threshold on the cosine similarity distribution to identify samples that belong to the same hierarchical grouping. However, the cosine similarity distributions generated for each SSL method will differ from each other and thus for some methods a specific type of thresholding may act as a better approximation for hierarchical groupings. However, it should be acknowledged that performance improves over baseline SSL approaches regardless of the type of decomposition. Further analysis of the threshold hyperparameter can be found in Section \ref{sec: additional_analysis}.

\begin{table}[]
\centering
\resizebox{\linewidth}{!}{
\begin{tabular}{ccccc}
             & \multicolumn{2}{c}{Linear}      & \multicolumn{2}{c}{Semi-Supervised} \\ \cline{2-5} 
Method       & Top-1          & Top-5          & 1\%              & 10\%             \\ \hline
SimCLR       & 57.74          & 81.44          & 29.61            & 47.85            \\
SimCLR + HEX & \textbf{59.03} & \textbf{82.83} & \textbf{33.72}   & \textbf{50.48}   \\ \hline
NNCLR        & 62.47          & 85.13          & 39.94            & 57.83            \\
NNCLR + HEX  & \textbf{65.96} & \textbf{87.02} & \textbf{46.00}   & \textbf{61.45}   \\ \hline
\end{tabular}}
\caption{This table shows the performance of HEX on the ImageNet dataset within a linear evaluation setting as well as within the context of semi-supervised fine-tuning on 1\% and 10\% subsets of ImageNet.}
\label{tab: imagenet_complete}
\end{table}

\begin{figure}[ht]
\centering
\includegraphics[scale = .4]{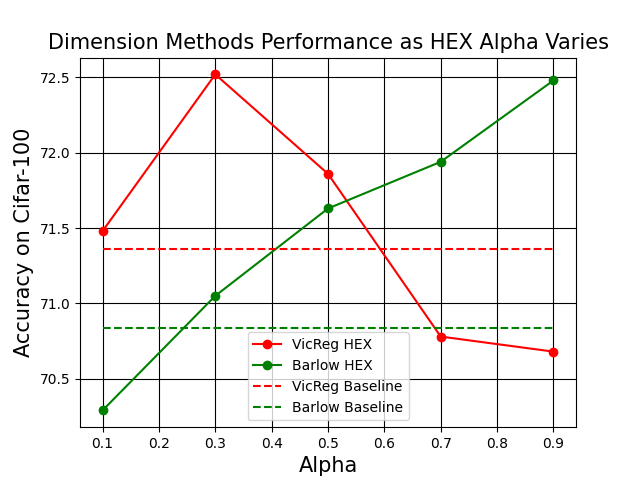}
\caption{This shows the effect of varying the alpha parameter when using HEX as regularization on top of a dimension contrastive approach. We use the adaptive decomposition strategy for this plot.}
\label{fig: dimension}
\end{figure}

\begin{table}[t]
\centering
\caption{This shows the performance of HEX on top of NNCLR pretrained with Imagenet within both a fine-grained fine-tuning setting on top of frozen representations as well as within a traditional transfer learning setting. The metric is top-1 accuracy.}
\resizebox{\linewidth}{!}{
\begin{tabular}{llcc|cc}

\toprule
\multicolumn{2}{c}{\textbf{Fine-Grained Dataset}} & \multicolumn{2}{c}{\textbf{Fine-Tuning}} & \multicolumn{2}{c}{\textbf{Transfer Learning}} \\ \cmidrule(r){3-4} \cmidrule(r){5-6}
 & & \textbf{NNCLR} & \textbf{NNCLR + HEX} & \textbf{NNCLR} & \textbf{NNCLR + HEX} \\ \midrule
Cars \cite{dehghan2017view} & & 45.36 & \textbf{56.83} & 88.39 & \textbf{88.45} \\
Flowers \cite{nilsback2008automated} & & 86.18 & \textbf{90.88} & 89.02 & \textbf{92.84} \\
NABirds \cite{van2015building} & & 33.09 & \textbf{44.58} & 65.75 & \textbf{68.06} \\
Food \cite{bossard2014food} & & 71.27 & \textbf{72.52} & 86.15 & \textbf{86.29} \\
Caltech-101 \cite{fei2004learning} & & \textbf{95.76} & 95.48 & 93.50 & \textbf{96.33} \\
Dogs \cite{KhoslaYaoJayadevaprakashFeiFei_FGVC2011} & & 64.10 & \textbf{70.50} & 71.80 & \textbf{73.22} \\
Pets \cite{parkhi2012cats} & & 82.15 & \textbf{86.84} & 82.15 & \textbf{86.84} \\
Aircraft \cite{maji13fine-grained} & & 44.61 & \textbf{53.19} & 83.65 & 83.65 \\
Caltech-Birds \cite{wah2011caltech} & & 44.96 & \textbf{57.33} & 69.83 & \textbf{72.66} \\ \bottomrule
\end{tabular}
}
\label{tab: transfer}
\end{table}

\begin{table}[]
\centering
\resizebox{\linewidth}{!}{
\begin{tabular}{cccccc}
Study                         &            & SimCLR         & SimCLR + HEX   & NNCLR & NNCLR + HEX    \\ \hline
\multirow{3}{*}{Clustering}   & K=1        & 50.13          & \textbf{50.86} & 60.08 & \textbf{60.45} \\
                              & K=5        & 54.96          & \textbf{55.23} & 61.46 & \textbf{63.62} \\
                              & K=10       & 56.85          & \textbf{57.13} & 64.31 & \textbf{64.6}  \\ \hline
\multirow{5}{*}{Transformers} & ViT\_tiny \cite{alexey2020image} & 49.61          & \textbf{51.68} & 32.95 & \textbf{35.75} \\
                              & ViT\_small \cite{alexey2020image}& 55.19          & \textbf{58.57} & 35.78 & \textbf{37.38} \\
                              & ViT\_base \cite{alexey2020image} & 57.87          & \textbf{60.1}  & 38.33 & \textbf{39.34} \\
                              & ViT\_large \cite{alexey2020image}& \textbf{28.05} & 27.83          & 37.47 & \textbf{43.17} \\
                              & Swin \cite{liu2021swin}       & 49.46          & \textbf{51.35} & 26.86 & \textbf{31.77} \\ \hline
\end{tabular}}
\caption{In this table, we show ablation studies related to using KNN clustering accuracy as a metric from models trained with adaptive thresholding as well as the performance when the backbone is swapped with different types of vision transformers. These experiments were done on Cifar-100 for the SimCLR and NNCLR methodologies and all backbone training was done using the same hyperparameters as the ResNet-50.}
\label{tab: ablation}
\end{table}

\begin{table}[]
\centering
\resizebox{\linewidth}{!}{
\begin{tabular}{ccccc}
\hline
                & SimCLR & SimCLR + HEX   & NNCLR  & \multicolumn{1}{l}{NNCLR + HEX} \\ \hline
iNat21          & 20.67  & \textbf{22.01} & 18.786 & \textbf{19.37}                  \\
TinyImageNet200 & 40.92  & \textbf{45.35} & 39.69  & \textbf{42.91}                  \\
STL-10          & 86.16  & \textbf{88.83} & 86.80  & \textbf{87.20}                  \\
Cifar-10        & 89.17  & \textbf{90.59} & 92.04  & \textbf{92.13}                  \\ \hline
\end{tabular}}
\caption{Performance of HEX on top of SimCLR and NNCLR across a wide variety of datasets of varying complexity in a linear evaluation setting. The adaptive thresholding strategy is used for HEX in this table.}
\label{tab: dataset}
\end{table}

We also note that the performance improvements of our method have a greater margin on Cifar-100 compared to Imagenet-100. Cifar-100 has more classes that share features in common with each other due to the intentional creation of superclass groupings in the dataset. This means that there is a greater potential for local collapse within these hierarchical groupings as compared to a dataset like Imagenet-100 where most classes are much more separable with respect to each other. This indicates that the HEX algorithm may help to a greater degree when there exists a greater degree of class overlap within the construction of the dataset. This perspective is further validated with our results on the Imagenet dataset \cite{deng2009imagenet} in Table \ref{tab: imagenet_complete}.  We see a significant improvement when using the HEX algorithm for both strategies. This improvement is greater compared to the Imagenet-100 subset as there is naturally more class overlap with the introduction of a larger number of classes from the Imagenet dataset. Therefore, a greater need to regularize against local collapse is necessary. This performance improvement is also observed in Table \ref{tab: imagenet_complete} where we show the results of semi-supervised fine-tuning of our frozen Imagenet representations on 1\% and 10\% of the available labels within the Imagenet dataset. Additionally, in Table \ref{tab: transfer}, we take the NNCLR models pre-trained with and without HEX on Imagenet and see how well its representations can be used in both fine-tuning with a frozen backbone as well as transfer learning scenarios for a wide variety of fine-grained datasets. We see that in the vast majority of cases the representations from the model that received additional HEX regularization results in a significant improvement in top-1 classification accuracy. This is significant because these fine-grained datasets exhibit a high degree of class overlap and thus may be more prone to local collapse of their representations than standard datasets. 

Another major contribution of our work is that our HEX loss can be used as an additional regularization on top of dimensional contrastive approaches like VicReg and Barlow Twins. We do this through the introduction of an $\alpha$ parameter weighting with the associated $L_{dim}$ for each method and show an explicit analysis of this in Figure \ref{fig: dimension}. We observe that by appropriately choosing the proper $\alpha$ parameter we can expect significant performance improvements over the baseline methods. This shows that dimensional contrastive strategies also have a hierarchical emergence characteristic during training and hence benefit from additional HEX regularization. This is especially interesting given that their optimization objective does not explicitly enforce sample-wise relationships. We also see that the appropriate choice of the $\alpha$ parameter varies based on the specific dimension-based approach. In this case, Barlow Twins improves almost linearly as the $\alpha$ parameter is increased while VicReg requires an equal or lesser weighting with the HEX loss. This may indicate that the degree of localized collapse varies for each approach and thus requires a differing degree of additional HEX regularization for optimal performance improvements. 

\subsection{Ablation Studies} 

We also test the robustness of our HEX algorithm under different metrics and backbones in Table \ref{tab: ablation}. We see that the performance improvements of HEX are maintained when using clustering accuracy to measure performance under different K-values in the KNN algorithm. Additionally, we replace the ResNet-50 backbone of our baseline experiments with different vision transformer backbones. We observe that the additional HEX algorithms improves performance with different sized vision transformers (ViT\_tiny to ViT\_large) and different families of transformers (Swin and ViT) thus demonstrating the wide applicability of our approach. Note that the performance difference between SimCLR and NNCLR may be due to the specific hyperparameters chosen for each method for our baseline experiments may be better or worse for specific transformer architectures. We also show performance improvements across datasets with varying complexity in Table \ref{tab: dataset}. Specifically, the high complexity datasets are iNat21 and TinyImageNet-200 while the lower complexity datasets are STL-10 and Cifar-10. In both cases, HEX results in significant performance improvements. This demonstrates the adaptability of our approach even when the exact hierarchical structure of the dataset is unclear or difficult to extract.

\section{Conclusion}
\label{sec: conclusion}

In this work, we propose an algorithm to take advantage of the emergence of hierarchical structures during the training process of self-supervised algorithms. We show through our analysis that the cosine similarity metric can effectively separate hierarchically grouped samples from regular ones. Therefore, by using an appropriate threshold parameter, we can effectively decompose the denominator of the InfoNCE loss to account for these local hierarchical structures. In this way, we are able to introduce an additional localized regularization that can counter the local collapse of samples within the same hierarchical grouping. We show that by appropriately integrating HEX on top of both dimension and sample contrastive SSL algorithms we can achieve significant performance improvements over a wide variety of datasets and training settings.

{\small
\bibliographystyle{ieee_fullname}
\bibliography{egbib}
}

\clearpage
\section{Supplementary}
\label{sec: supplementary}


\subsection{Dataset Details}
We evaluate the performance of HEX across different domains and levels of granularity. To assess the robustness and versatility of HEX, we choose 15 diverse datasets that span a wide range of scenarios that vary based on size, diversity, granularity, and difficulty of the data setting. These datasets include popular ones for benchmarking such as CIFAR-100 \cite{krizhevsky2009learning}, Cifar-10   \cite{krizhevsky2009learning}, TinyImageNet-200 \cite{yao2015tiny}, STL-10     \cite{coates2011analysis}, iNaturalist 2021 \cite{van2018inaturalist}, and ImageNet \cite{deng2009imagenet}. Additionally, our intent is to also conduct experiments under conditions with significant class similarity, which makes them more susceptible to local collapse in their representations compared to standard datasets. Therefore, we select nine fine-grained datasets including FGVC Aircraft \cite{maji13fine-grained}, Oxford 102 Flowers \cite{nilsback2008automated}, Stanford Cars \cite{dehghan2017view}, Stanford Dogs \cite{KhoslaYaoJayadevaprakashFeiFei_FGVC2011}, Oxford Pets \cite{parkhi2012cats}, NABirds  \cite{van2015building}, Caltech 101 \cite{fei2004learning}, Food 101 \cite{bossard2014food}, and CUB-200-2011 \cite{wah2011caltech}. A description of each dataset is given in Table \ref{tab:dataset_configuration}.

In addition to ImageNet, we include its subset, ImageNet-100 \cite{deng2009imagenet}, for comparative experiments. ImageNet-100 introduces a smaller number of classes than ImageNet and naturally has less class overlap. By comparing datasets with similar data distribution conditions, but differing only in the degree of class overlap, we aim to observe how our method improves performance by preventing local collapse.

\subsection{Limitations of Algorithm}

Despite the advantages of HEX, there are some potential limitations. For example, in certain cases manual choosing the thresholding parameter performs better than the adaptive strategy. Consequently, some type of hyperparameter search would have to occur in these situations. Additionally, our method seems to perform better in situations with a higher degree of class overlap and complexity. Therefore, performance benefits may not be as great within lower complexity data settings. Furthermore, our algorithm relies on the emergence of hierarchical structures during training. However, it isn't always clear from a human perspective what the nature of a hierarchy can look like for certain application domains. Further research into the nature of hierarchies in certain domains may be needed to fully understand the meaning of hierarchical emergence in general.

\subsection{Additional Training Details}

\subsubsection{Code Acknowledgement}
We make use of the solo-learn codebase for all experiments \cite{da2022solo}. The link to their code can be found at this \href{https://github.com/vturrisi/solo-learn}{repository}. Our specific implementations will be released upon acceptance of the paper.

\subsubsection{Specific SSL Method Details} In Table \ref{tab:details}, we show all the hyperparameters associated with each of the SSL algorithms highlighted in the results section of our paper. The hyperparameters in the table are the basic hyperparameters that were associated with the solo-learn codebase. However, these parameters change slightly depending on the applied dataset. These changes were minor and very method-specific. Examples of this include slight variations in the queue size to reflect the size of the dataset that the SSL method was trained on. Otherwise, this table reflects the basic training hyperparameters for all SSL methods in the paper.

Additionally, there were instances within the paper where integration of the HEX loss required structural changes to the optimization process of the associated SSL method. For example, the losses for SimCLR and NNCLR were directly replaced by their HEX version of the loss. In Section \ref{sec: methodology}, we detail the version of the loss that directly replaces the SimCLR loss. The NNCLR version of HEX is equivalent to the SimCLR version of HEX except the augmented sample in the positive set is that of its nearest neighbor, rather than derived from the same sample as the anchor image. The All4One methodology uses the NNCLR loss as part of their methodology. However, the HEX version of All4One replaces its NNCLR loss with the NNCLR + HEX version. In the case of dimensional contrastive approaches like VicReg and Barlow Twins, we add the SimCLR version of HEX to each of these losses with an $\alpha$ parameter to weight the contribution of each loss. For VicReg specifically, we also had to consider the hyperparameters that went into weighting each of its other loss functions. We empirically found that scaling the HEX loss by a factor of 5 allowed stable training alongside the other loss functions in the VicReg formulation. Barlow Twins did not require any additional tuning to adapt to additional HEX regularization.

All methods received the default augmentation policy described within the solo-learn github. The only parameter that was changed in a dataset specific manner was the random resized crop parameter. If the images were smaller than 224x224, they received a random crop size that is equal to the size of the images within the dataset. However, if the images were larger, then the parameter was defaulted to 224x224. 

\subsubsection{Transformer Experiments}
We used the standard vision transformer and the Swin transformer for all experiments in our study. We kept all hyperparamters and optimizers the same as in the case of the ResNet-50 experiments with the exception of transformer specific changes such as a decoder embed dimension of 512, depth of 8, patch size of 4, and 16 decoder heads.

\subsubsection{Hyperparameter Variation} 
\begin{figure}[ht]
\centering
\includegraphics[scale = .45]{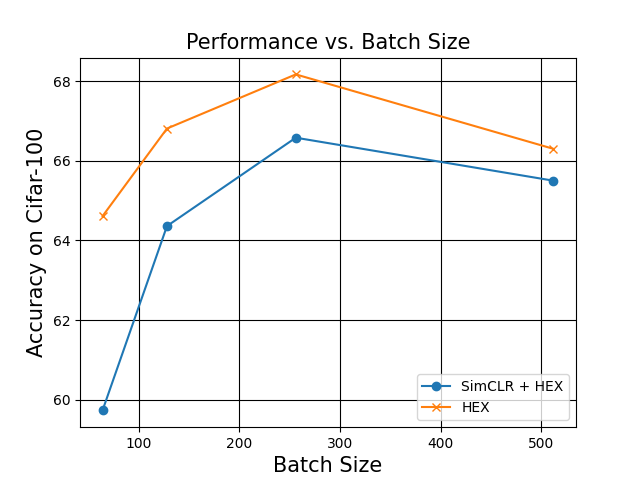}
\caption{This shows performance variation as batch size is varied. The HEX approach is able to maintain performance improvements.}
\label{fig: batch}
\end{figure}

We show in Figure \ref{fig: batch} the impact of varying batch size on the performance of our method. We see that the performance improvements of HEX are maintained even when this hyperparameter is varied.

\label{sec: train_details}
\begin{table*}[]
\centering
\begin{tabular}{cccccccc}
\hline
Method  & Projection        & \begin{tabular}[c]{@{}c@{}}Method Specific \\ Parameters\end{tabular}                         & Temperature & Optimizer & Batch Size & LR  & Decay \\ \hline
SimCLR  & 2048-128          & N/A                                                                                           & 0.1         & LARS      & 256        & 0.4 & 1e-5  \\
NNCLR   & 2048-4096-256     & Queue = 98304                                                                                 & 0.2         & LARS      & 256        & 1.0 & 1e-5  \\
All4One & 2048-4096-256     & \begin{tabular}[c]{@{}c@{}}Queue = 98304\\ Momentum = 0.99\end{tabular}                       & 0.2         & LARS      & 256        & 1.0 & 1e-5  \\
Barlow  & 2048 - 2048       & Scale Loss = 0.1                                                                              & N/A         & LARS      & 256        & 0.3 & 1e-4  \\
BYOL    & 4096 - 256 - 4096 & Momentum = 0.99                                                                               & N/A         & LARS      & 256        & 1.0 & 1e-5  \\
Moco v2 & 2048-256          & \begin{tabular}[c]{@{}c@{}}Queue Size = 65536\\ Momentum = 0.99\end{tabular}                  & 0.2         & SGD       & 256        & 0.3 & 1e-4  \\
VicReg  & 2048 - 2048       & \begin{tabular}[c]{@{}c@{}}Sim\_weight = 25\\ Var\_weight = 25\\ Cov\_weight = 1\end{tabular} & N/A         & LARS      & 256        & 0.3 & 1e-4  \\
SimSiam & 2048-2048-512     & N/A                                                                                           & 0.2         & SGD       & 256        & 0.5 & 1e-5  \\ \hline
\end{tabular}
\caption{This table shows the baseline training parameters for each SSL method. Variations from these parameters are discussed in the text.}
\label{tab:details}
\end{table*}

We also show an analysis of varying the threshold hyperparameter when using manual strategies in Table \ref{tab:hyperparameter_analysis}. We see that the performance on the CIFAR-100 dataset is sensitive to the chosen thresholding parameters. This demonstrates the superiority of the adaptive threshold setting method over the manual threshold setting method as this does not require potentially expensive tuning of hyperparameters. In Table \ref{tab:hyperparameter_analysis}, we also detail how these hyperparameters were manually set. This, alongside improved results, illustrates that manual hyperparameter tuning is potentially a sub-optimal strategy to the adaptive method when considering the results across all experiments on the CIFAR-100 dataset. In the Table, ``cos" refers to an additional thresholding strategy that lowers the threshold value in a continuous fashion from its starting value to a minimal value along a cosine curve that is a function of the epoch of training.

\begin{table}[h!]
\centering
\footnotesize
\begin{tabularx}{\textwidth}{cccc|c|c}
\multicolumn{4}{c|}{\textbf{Hyperparameter Tuning of Manual Strategy}} & \multicolumn{2}{c}{\textbf{Accuracy}} \\
dist\_thres & dist\_min & step\_down & step\_type & Top-1 & Top-5 \\

\cline{1-6}
0.95 & 0.65 & - & cos & 66.63 & 88.96 \\
0.85 & 0.45 & - & cos & 60.57 & 86.37 \\
0.95 & 0.65 & 0.1 & step & 66.20 & 89.04 \\
0.85 & 0.45 & 0.1 & step & 65.63 & 88.18 \\
0.95 & 0.65 & 0.05 & step & 66.23 & 88.70 \\
0.85 & 0.45 & 0.05 & step & 65.08 & 88.29 \\

\cline{1-6}
\multicolumn{4}{c|}{\textbf{Proposed Adaptive Strategies}} & \textbf{67.56} & \textbf{90.02} \\
\end{tabularx}
\caption{This demonstrates 6 different manual hyperparameter tuning approaches. In CIFAR-100 datasets, we can see none of them were superior to the adaptive strategy. ``step" refers to the stepwise manual strategy while ``cos" represents the cosine thresholding strategy.}
\label{tab:hyperparameter_analysis}
\end{table}

\subsection{Additional Analysis}
\label{sec: additional_analysis}

\begin{figure}[ht]
\centering
\includegraphics[scale = .45]{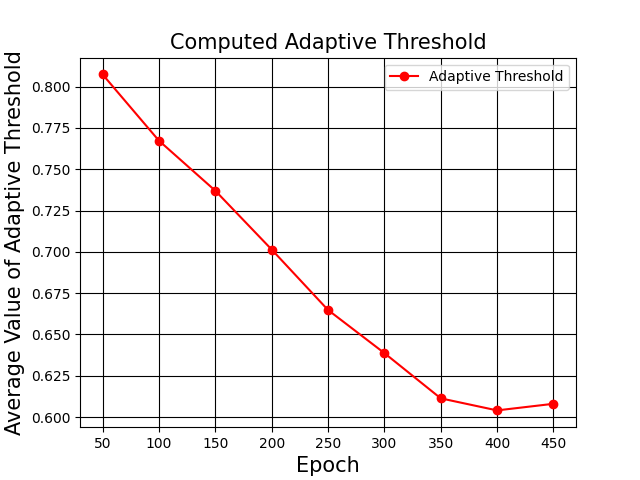}
\caption{This shows the value of the threshold parameter in the adaptive setting across different epochs of training on the CIFAR-100 test set.}
\label{fig: ada}
\end{figure}

\subsubsection{Threshold Analysis} The adaptive version of the HEX loss uses the cosine similarity distribution to assign a threshold at different points in training. We show an example of the exact value of this threshold at different points in training in Figure \ref{fig: ada}. This was computed as the average threshold value of all 10000 images in the CIFAR-100 test set. However, in practice this threshold is computed on a batch-wise basis rather than across the entire dataset as a whole.  We see that the adaptive threshold gradually decreases during training to reflect the shifting cosine similarity distribution of the dataset. Later in training, this decrease starts to level out as the hierarchical structure of the dataset emerges with persistent larger cosine similarity values at samples within the same hierarchical grouping. 

\begin{figure}[ht]
\centering
\includegraphics[scale = .45]{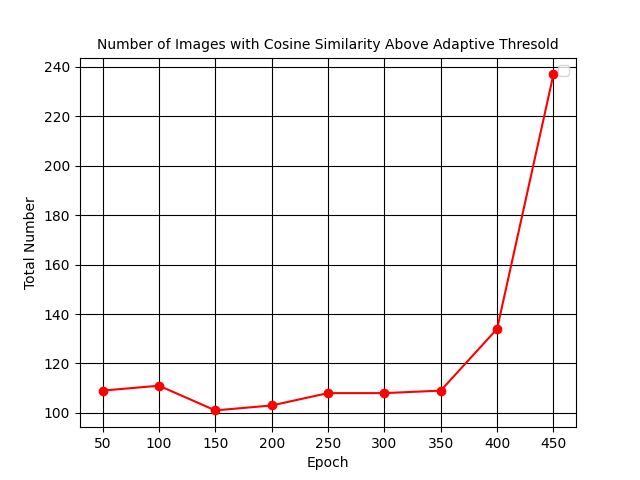}
\caption{This shows the value of the threshold parameter in the adaptive setting across different epochs of training on the Cifar-100 test set.}
\label{fig: total}
\end{figure}

Additionally, in Figure \ref{fig: total} we show the average number of samples in the Cifar-100 test set are above the threshold parameter at different epochs of training. This value gradually increases as the threshold parameter is lowered as a result of growing confidence in the hierarchical structure of the representation space at later points of training.


\subsubsection{Re-weighting Ablation} We also show an ablation study comparing our method against a different approach \cite{robinson2020contrastive} that also introduces a re-weighting of negative terms. The difference is that our method explicitly identifies samples at different points of training based on the intuition of identifying hierarchical groupings while the other approach generally re-weights the entire batch as a whole. We show performance improvements in Table \ref{tab:comparison} within a fine-tuning setting on Cifar-100 with the training parameters described in this paper while using a stepwise strategy for HEX. This shows that our performance improvements are not just a result of reweighting, but also due to the importance of identifying the right hierarchically aligned samples. 
\begin{table}[]
\centering
\begin{tabular}{cc}
\hline
Method & Cifar-100 Linear Evaluation Accuracy \\ \hline
HCL    & 64.64                          \\
SimCLR & 64.46                          \\
HEX    & 67.56                         \\ \hline
\end{tabular}
\caption{We show a comparison with a different re-weighting strategy that does not have a mechanism to target hierarchical groupings.  HCL is trained in the manner described in its paper, \cite{robinson2020contrastive} while we train SimCLR and HEX in the manner described in this paper.}
\label{tab:comparison}
\end{table}

\subsubsection{Fine-grained pretraining Ablation}
We show an ablation study to evaluate the efficacy of our proposed HEX method on a less diverse representation space. Fine-grained datasets are less diverse due to all their classes sharing features in common with each other. In the main paper, we showed the result of fine-tuning and transfer learning within the context of a pre-trained Imagenet model. Instead of pretraining on ImageNet datasets, which has greater feature diversity, we pre-train a NNCLR model with fine-grained datasets as depicted in Table \ref{tab:pretrain_finegrained}. We then compare against the same NNCLR setup with our additional HEX regularization.
As shown, applying HEX regularization during SSL pretraining consistently results in better classification performance across most datasets compared to not applying HEX. This demonstrates that our HEX technique effectively disentangles the local-collapse phenomena.

\begin{table}[h!]
\centering
\begin{tabular}{lcc|cc}
\hline
 & \multicolumn{2}{c|}{NNCLR} & \multicolumn{2}{c}{NNCLR + HEX} \\
 & Top-1 & Top-5 & Top-1 & Top-5 \\
\hline
Cars \cite{dehghan2017view} & 42.92 & 71.99 & \textbf{46.60} & \textbf{75.66} \\
Flowers \cite{nilsback2008automated} & \textbf{32.94} & \textbf{59.22} & 28.63 & 55.88 \\
NABirds \cite{van2015building} & 21.83 & 45.19 & \textbf{22.26} & \textbf{45.67} \\
Caltech-101 \cite{fei2004learning} & 72.60 & 93.79 & \textbf{73.73} & \textbf{94.35} \\
Dogs \cite{KhoslaYaoJayadevaprakashFeiFei_FGVC2011} & 42.06 & 73.75 & \textbf{47.40} & \textbf{77.95} \\
Pets \cite{parkhi2012cats} & 47.78 & 81.90 & \textbf{53.23} & \textbf{85.88} \\
Aircraft \cite{maji13fine-grained} & \textbf{30.03} & \textbf{57.82} & 29.07 & 56.68 \\
Caltech-Birds \cite{wah2011caltech} & 19.49 & 43.22 & \textbf{20.76} & \textbf{47.36} \\
\hline
\end{tabular}
\caption{By comparing with and without HEX methods on pretrained models with fine-grained datasets, we show that HEX can still enhance classification performance in a less diverse representation space.}
\label{tab:pretrain_finegrained}
\end{table}

\begin{table}[]
\centering
\begin{tabular}{clcl}
\multicolumn{1}{l}{} &      & \multicolumn{2}{c}{iNat21}      \\ \hline
                     & Type & Top-1          & Top-5          \\ \hline
SimCLR               & None & 20.67          & 38.64          \\
SimCLR + HEX         & Ada  & 22.01          & 40.92          \\
SimCLR + HEX         & Sup  & \textbf{23.35} & \textbf{42.94} \\ \hline
\end{tabular}
\caption{This shows the performance of HEX using one of the superclasses within iNat21.}
\label{tab: iNat21_super}
\end{table}
\begin{table}[]
\centering
\begin{tabular}{clcl}
\multicolumn{1}{l}{} &      & \multicolumn{2}{c}{Imagenet}      \\ \hline
                     & Type & Top-1          & Top-5          \\ \hline
All4One               & None & 52.83         & 78.29          \\
All4One  + HEX         & Ada  & 54.30         & 79.09         \\
 \hline
\end{tabular}
\caption{This shows the performance of HEX on All4One on the Imagenet dataset with 50 epochs of pre-training within the linear evaluation setting.}
\label{tab: all4one}
\end{table}

\subsubsection{Hierarchical Emergence}

\begin{figure}[ht]
\centering
\includegraphics[scale = .2]{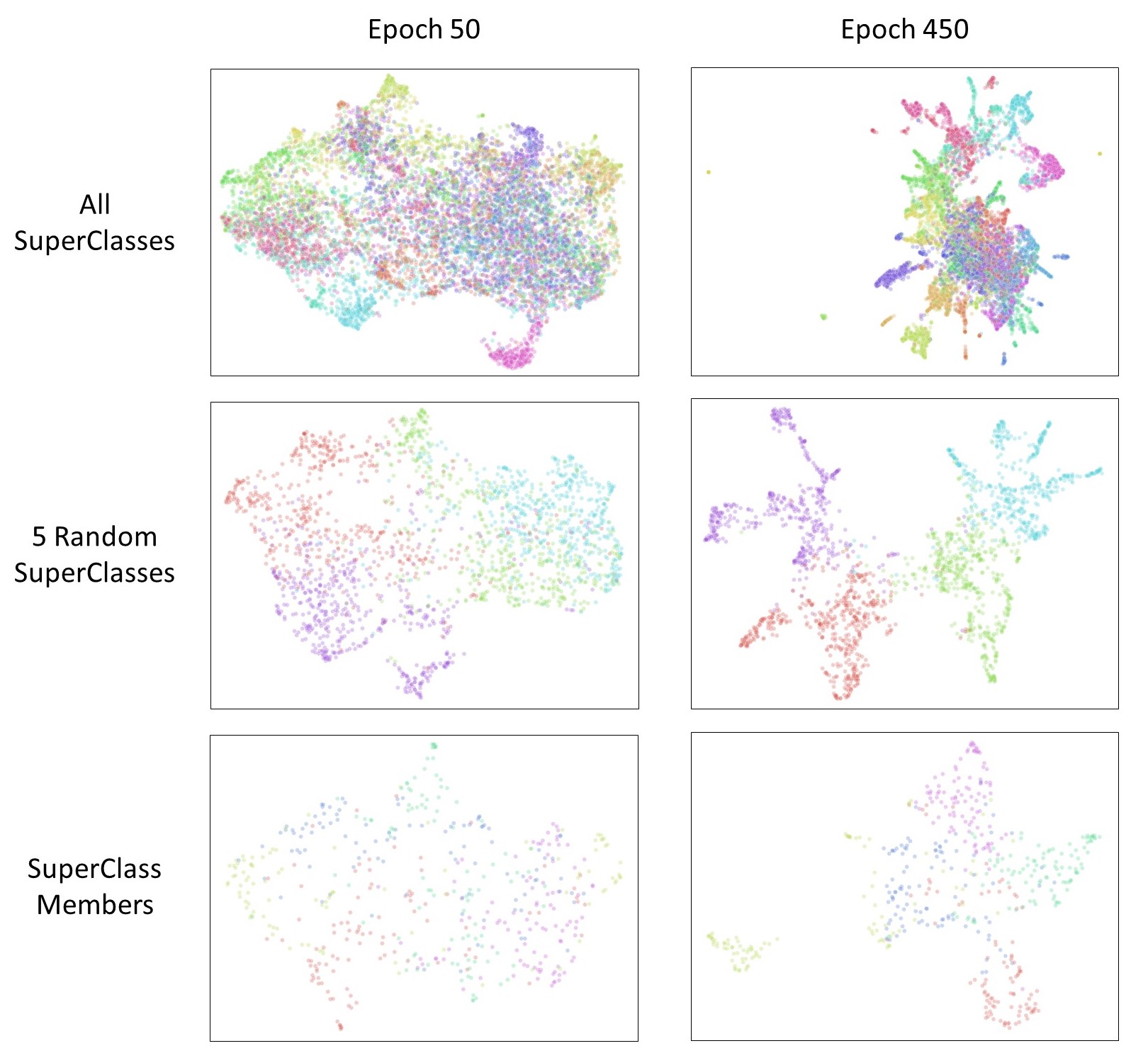}
\caption{This plot shows the umap embeddings of the samples of Cifar-100 labeled by their superclass label. We show the organization of the representation space under different data access settings.}
\label{fig: umap}
\end{figure}

\begin{figure}[ht]
\centering
\includegraphics[scale = .5]{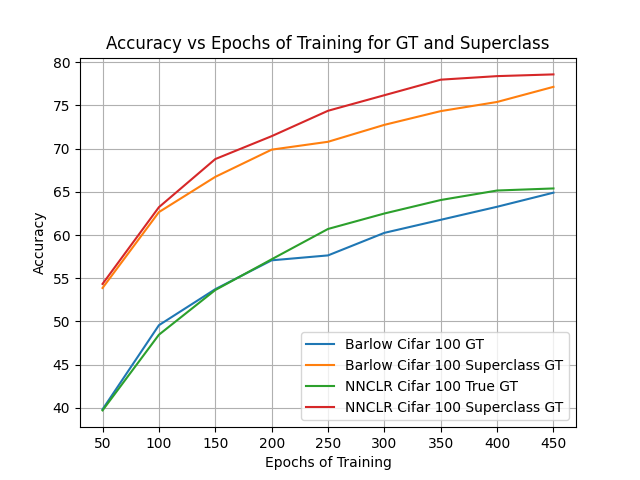}
\caption{This shows the KNN accuracy using the cosine similarity distance metric for the NNCLR \cite{dwibedi2021little} and Barlow Twins \cite{zbontar2021barlow} SSL methods across 450 epochs of training for both ground truth Cifar-100 labels as well as labels denoting the superclass of each sample. }
\label{fig: knn}
\end{figure}

 To get an intuitive sense of the emergence of hierarchies during training, we take each model $m_e$ described in Section \ref{sec: hierarchical_analysis} and pass in the test set to get an associated representation matrix $R$ with each row corresponding to a test image representation $r_i$. We associate each $r_i$ with its label $y_i$ and superclass label $y_{si}$. Together, this information is used to produce Figures  \ref{fig: umap} and \ref{fig: knn}. In Figure \ref{fig: umap} we show a UMAP \cite{mcinnes2018umap} visualization of the representation space at the beginning and end of training under different conditions and subsets of the test data space. In the top row of this figure, we see all points in the test set labeled by their superclass. It is visually evident that all superclass clusters become more separable by the end of training. This is shown more clearly in the middle row of Figure \ref{fig: umap} where we randomly choose samples from 5 random superclasses and show the organization of their clusters at the beginning and end of training. We also analyze the organization of classes within a single superclass in the bottom row of Figure \ref{fig: umap}. These plots show that classes within a superclass also become more separable with respect to each other. All of these plots together indicate that representations naturally cluster in terms of superclasses as well as regular ground truth classes over the course of SSL training. This is further evidenced in Figure \ref{fig: knn} where we observe that the KNN accuracy of the representations with respect to both superclass and ground truth labels both increase over the course of training for the dimension-based Barlow Twins \cite{zbontar2021barlow} and sample-based NNCLR strategies. 

 We also show an additional study of using hierarchies in the iNaturalist21 dataset. This dataset is composed in a hierarchical fashion that corresponds to the taxonomies that come with animal and plant species. We show in Table \ref{tab: iNat21_super} that appropriately using the hierarchy information for HEX in a supervised fashion of the associated superclasses leads to performance improvements in the same way as the Cifar-100 datasets in the main paper. 

 \subsubsection{Additional ImageNet Experiments}
We also added an additional experiment with 50 epochs of pre-training with the All4One methodology. This is shown in Table \ref{tab: all4one}. We see that HEX is able to achieve performance improvements on the Imagenet dataset for this algorithm as well.

\subsubsection{Embedding Space Analysis}

\begin{figure}[ht]
\centering
\includegraphics[scale = .5]{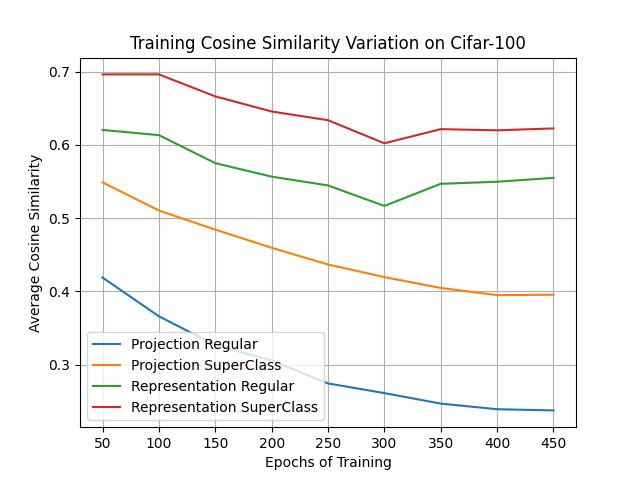}
\caption{This shows the variation in average pairwise cosine similarity between superclass negatives and regular negatives for each instance in the Cifar-100 test set throughout different epochs of training. This plot is from the NNCLR method.}
\label{fig: cosine}
\end{figure}

\begin{figure}[ht]
\centering
\includegraphics[scale = .5]{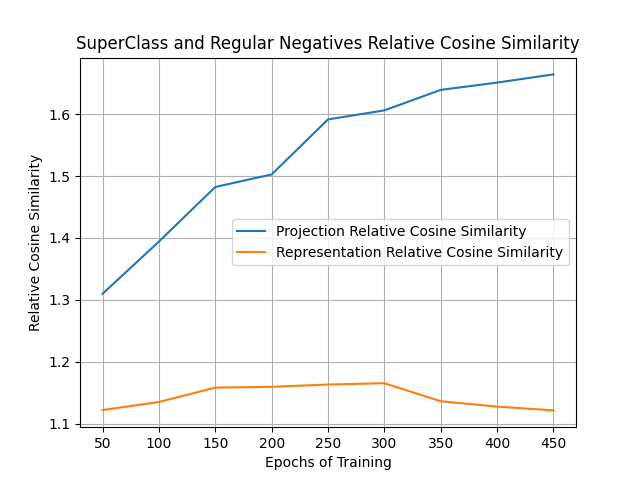}
\caption{This shows the relative cosine similarity between superclass and regular negatives over the course of training on Cifar-100 with the NNCLR method.}
\label{fig: rel_cosine}
\end{figure}

We analyze a variety of trends with respect to cosine similarity distributions in Figures \ref{fig: cosine} and \ref{fig: rel_cosine}. In Figure \ref{fig: cosine}, we show the average cosine similarity value for each anchor image with all other images in the test set of Cifar-100 at different epochs of training with the NNCLR method within both the projection space and representation space of the model. On average, for both spaces the average cosine similarity of superclass samples is higher than that of regular samples. Additionally, we see that the average cosine similarity of the representation space is higher than that of the projection space possibly due to retaining a greater number of redundant features between samples. From this analysis, it is unclear whether to compute the cosine similarity with respect to the representation space or projection space when estimating the presence of hierarchical groupings. To analyze these trends further, we take the ratio of superclass cosine similarity to regular class cosine similarity for both spaces across all epochs of training. We observe in Figure \ref{fig: rel_cosine} that this ratio increases over training for the projection space, but not the representation space. This indicates that the cosine similarity metric of the embedding space becomes more hierarchically aligned as training progresses due to a greater relative difference in the cosine similarity between the two subsets. It is interesting to note that this does not hold for the cosine similarity of the representation space which shows that the space in which this metric is computed matters in terms of serving as a useful indicator of hierarchical separability. 

\begin{figure}[ht]
\centering
\includegraphics[scale = .3]{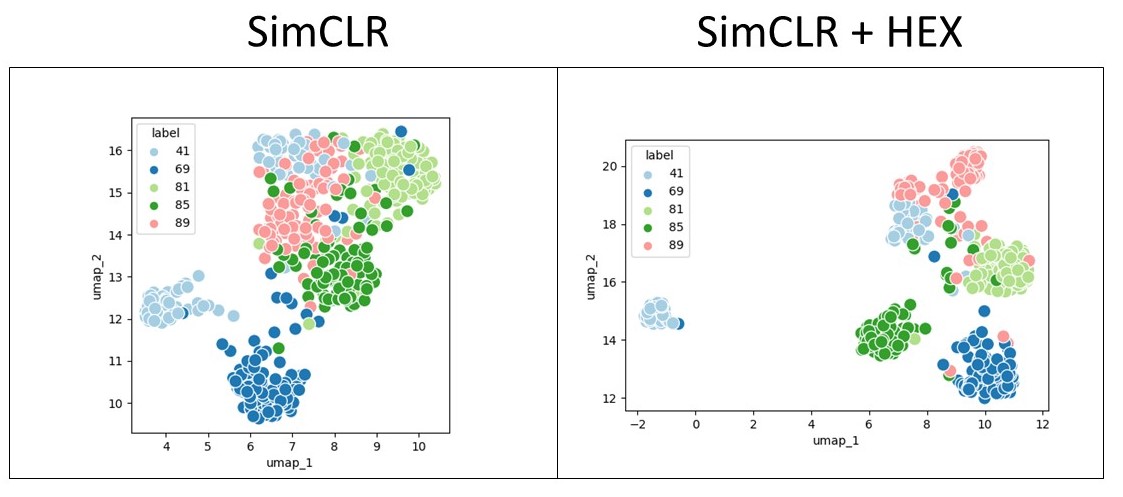}
\caption{This shows the embedding space of the vehicle superclass before and after the application of HEX to the SimCLR algorithm.}
\label{fig: hex_umap}
\end{figure}

We also show in Figure \ref{fig: hex_umap}, the embedding space of the Cifar-100 test set for samples from the vehicle superclass for both SimCLR and SimCLR + HEX. This visualization was created using the UMAP algorithm on the representation space of each model. We see that the application of HEX causes additional spread for these locally clustered samples within the same superclass. This results in greater separability between classes that are prone to collapsing with respect to each other compared to the embedding space produced by SimCLR alone. 

\begin{figure}[ht]
\centering
\includegraphics[scale = .5]{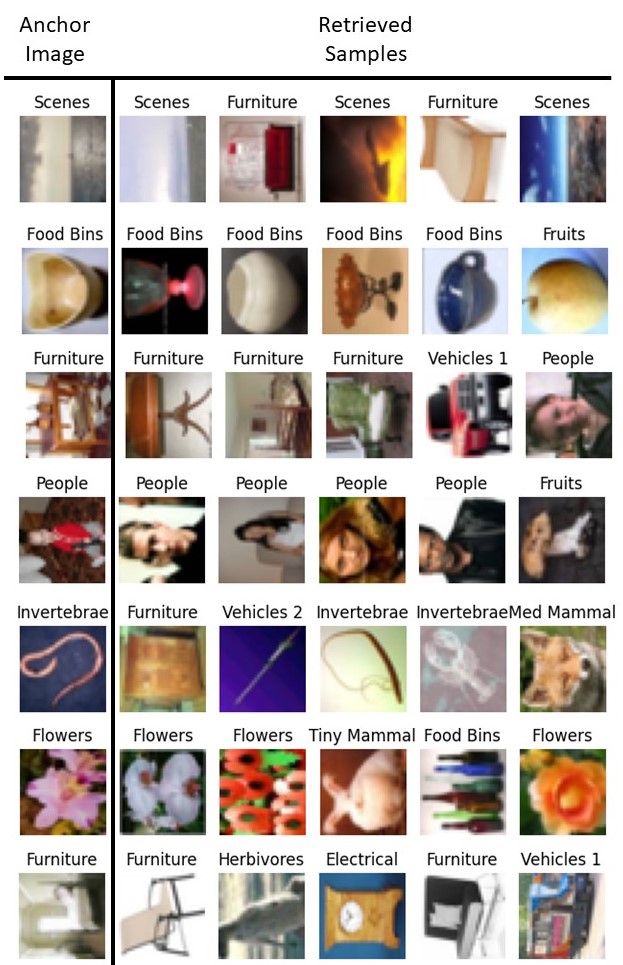}
\caption{This shows the images with the highest cosine similarity in each batch with the given anchor image. The label above each image identifies the superclass that it belongs to. These images were drawn from the Cifar-100 test set.}
\label{fig: retrieval}
\end{figure}

We also visually show the samples with the highest cosine similarity in randomly generated batches of 128 as each anchor image in Figure \ref{fig: retrieval}. This plot was produced using a model trained with the SimCLR methodology for 400 epochs of training. Every image is labeled with the superclass that it belongs to. Note that in most cases, the retrieved samples are members of the same superclass as each anchor image.

\clearpage
\begin{table*}[h!]
\centering
\begin{tabularx}{\textwidth}{ll|X|r}
Dataset & Abbreviation \& Link & Description & \# of classes \\
\hline
CIFAR-100 \cite{krizhevsky2009learning} & \href{https://www.cs.toronto.edu/~kriz/cifar.html}{cifar100} & 100 classes of 32x32 color images, including animals, vehicles, and various objects commonly found in the world. & 100 \\
\hline
CIFAR-10 \cite{krizhevsky2009learning} & \href{https://www.cs.toronto.edu/~kriz/cifar.html}{cifar10} & 10 classes of 32x32 color images featuring everyday objects and scenes such as airplanes, cars, and animals. & 10 \\
\hline
Tiny ImageNet \cite{yao2015tiny} & \href{http://tiny-imagenet.herokuapp.com/}{tinyimagenet200} & 200 classes of 64x64 images, a smaller version of the ImageNet dataset, used for object recognition and classification tasks. & 200 \\
\hline
STL-10 \cite{coates2011analysis} & \href{http://ai.stanford.edu/~acoates/stl10/}{stl10} & 10 classes of 96x96 images, designed for developing unsupervised feature learning, deep learning, and self-taught learning algorithms. & 10 \\
\hline
iNaturalist 2021 \cite{van2018inaturalist} & \href{https://www.kaggle.com/c/inaturalist-2021}{inat21} & Large-scale dataset with over 10,000 species, collected from photographs of plants and animals in their natural environments for fine-grained classification. & 10,000 \\
\hline
ImageNet \cite{deng2009imagenet} & \href{http://www.image-net.org/}{imagenet} & Large dataset with over 1,000 classes, used for image classification and object detection, containing millions of images across a wide variety of categories. & 1,000 \\
\hline
ImageNet-100 & \href{http://www.image-net.org/}{imagenet100} & Subset of ImageNet with 100 classes, providing a more manageable dataset for specific research and development purposes. & 100 \\
\hline
FGVC Aircraft \cite{maji13fine-grained} & \href{https://www.robots.ox.ac.uk/~vgg/data/fgvc-aircraft/}{aircraft} & Aircraft categorization dataset with 100 classes, featuring various aircraft models including different variants and manufacturers. & 100 \\
\hline
Oxford 102 Flowers \cite{nilsback2008automated} & \href{http://www.robots.ox.ac.uk/~vgg/data/flowers/102/}{flowers} & 102 category flower dataset, containing images of flowers commonly found in the United Kingdom, used for fine-grained visual classification. & 102 \\
\hline
Stanford Cars \cite{dehghan2017view} & \href{http://ai.stanford.edu/~jkrause/cars/car_dataset.html}{cars} & Car categorization dataset with 196 classes, covering a wide range of car models from various manufacturers, including different years and trims. & 196 \\
\hline
Stanford Dogs \cite{KhoslaYaoJayadevaprakashFeiFei_FGVC2011} & \href{http://vision.stanford.edu/aditya86/ImageNetDogs/}{dogs} & Dog breed classification dataset with 120 classes, containing images of different dog breeds, used for fine-grained classification tasks. & 120 \\
\hline
Oxford Pets \cite{parkhi2012cats} & \href{http://www.robots.ox.ac.uk/~vgg/data/pets/}{pets} & Cat and dog breeds dataset with 37 classes, each breed has roughly 200 images, used for pet recognition and classification. & 37 \\
\hline
NABirds \cite{van2015building} & \href{https://dl.allaboutbirds.org/nabirds}{NABirds} & North American birds dataset with 1011 species, featuring images of birds in various poses and environments, used for fine-grained bird species identification. & 1011 \\
\hline
Caltech 101 \cite{fei2004learning} & \href{http://www.vision.caltech.edu/Image_Datasets/Caltech101/}{caltech-101} & 101 object categories dataset and background, containing images of various objects including animals, buildings, and tools, used for object recognition and classification. & 102 \\
\hline
Food 101 \cite{bossard2014food} & \href{https://data.vision.ee.ethz.ch/cvl/datasets_extra/food-101/}{food} & 101 food categories with 101,000 images, featuring various dishes and cuisines from around the world, used for food recognition tasks. & 101 \\
\hline
CUB-200-2011 \cite{wah2011caltech} & \href{http://www.vision.caltech.edu/visipedia/CUB-200-2011.html}{Caltech-Birds} & 200 bird species dataset with annotated bounding boxes and part locations, used for fine-grained bird species classification and localization. & 200 \\
\hline
\end{tabularx}
\caption{Overview of various image datasets}
\label{tab:dataset_configuration}
\end{table*}
\clearpage

\end{document}